%% file: iclr2025_conference.tex
\documentclass{article} % For LaTeX2e
\usepackage{iclr2025_conference,times}

\input{math_commands.tex}

\usepackage{hyperref}
\usepackage{url}
\usepackage{subcaption}  % 用于并列子图
\usepackage{adjustbox} 
\usepackage{booktabs}
\usepackage{graphicx}
\usepackage{multirow}
\usepackage{colortbl}
\usepackage{diagbox}
\usepackage{titlesec}
\titlespacing{\section}{0pt}{*0.8}{*0.2}
\title{Advancing Multimodal In-Context Learning in Large Vision-Language Models with Task-aware Demonstrations}

\author{Yanshu Li \\
Brown University\\
Providence, RI 02912, USA \\
\texttt{yanshu\_li1@brown.edu}\\
}

\iclrfinalcopy
\begin{document}

\maketitle

\begin{abstract}
Multimodal in-context learning (ICL) has emerged as a key capability of Large Vision-Language Models (LVLMs), driven by their increasing scale and applicability. Despite its promise, effective ICL in the multimodal setting remains challenging due to the inherent complexity of image-text inputs and the high sensitivity of ICL performance to input configurations. In this work, we shed light on the core mechanism underlying multimodal ICL, identifying task mapping as a crucial factor in configuring robust in-context demonstration (ICD) sequences. Building on these insights, we propose \textit{SabER}, a lightweight yet powerful decoder-only transformer equipped with task-aware attention, which intelligently selects and arranges ICDs from a demonstration library in an autoregressive fashion. This design enables fine-grained feature extraction and cross-modal reasoning, iteratively refining task mapping to generate high-quality ICD sequences. Through extensive experiments covering five LVLMs and nine benchmark datasets, SabER not only demonstrates strong empirical performance, but also provides deeper understanding of how task semantics interact with multimodal ICDs. Our findings highlight the importance of principled ICD sequence configuration and open new avenues to enhance multimodal ICL in a wide range of real-world scenarios.
\end{abstract}

\section{Introduction}

As the demand for Large Language Models (LLMs) in real-world applications continues to surge, researchers have increasingly turned to prompt engineering and related techniques to enable these models to rapidly and accurately adapt to new tasks without the need for parameter updates \citep{brown2020lang, lester2021power, liu2021pretrain}. With the continual scaling of LLMs, a remarkable emergent property has been observed: the ability to perform complex reasoning and tackle novel tasks using only a handful of in-context demonstrations (ICDs) provided during a forward pass \citep{olsson2022incon, garg2023transformers}. This phenomenon, known as in-context learning (ICL), has fundamentally reshaped our understanding of task adaptation in modern LLMs.

The success of ICL in text-based settings has spurred efforts to extend its benefits to the multimodal domain. By incorporating interleaved image-text data into training corpora, Large Vision-Language Models (LVLMs) have naturally acquired robust multimodal ICL capabilities \citep{bai2023qwenvl, sun2024emu}. These models demonstrate promising potential in learning and reasoning from limited labeled data across various vision-language tasks—a particularly valuable trait given the challenges associated with assembling large-scale multimodal datasets \citep{ad48f017, tsimpoukelli2021mul}. However, as ICL moves beyond text to embrace more structured modalities, its performance becomes increasingly sensitive to the selection, order, and structure of ICD sequences \citep{10350488, zhou2024visual}. The complex interdependencies inherent in multimodal ICDs heighten the risks of modality misalignment and introduce task-irrelevant biases, thereby complicating the effective deployment of ICL in such settings.

Therefore, the configuration of ICD sequences holds even greater practical significance in multimodal ICL applications. However, research on this issue remains underexplored. Most existing studies on ICD sequence configuration focus solely on text matching and processing, making their direct adaptation to multimodal settings difficult \citep{iter2023in, 10448239}. Moreover, the underlying mechanisms of ICL in LVLMs are not yet well understood, despite being critical for designing effective ICD sequences. Unlike LLMs, where ICL primarily relies on implicit token retrieval, LVLMs must navigate intricate cross-modal interactions, making it unclear how they generalize patterns across different input formats. Without a principled understanding of how ICD sequences influence LVLM reasoning, current heuristic-based approaches to sequence configuration remain suboptimal, underscoring the need for a more systematic and mechanism-driven approach.

Our goal is to develop a more systematic understanding of LVLM's ICL and, based on this, design an end-to-end approach for achieving complete and high-quality ICD sequence configuration. First, we transfer the concepts of TR and TL to the multimodal domain and refine them for LVLMs. Using these insights, we introduce a new component, query, into traditional ICD configuration to improve modality coordination. We then systematically analyzed the roles of TR and TL in LVLMs based on this new configuration and found that task semantics is crucial for well-trained LVLMs. Building on this analysis, we propose \textit{SabER}, a novel tiny language model that optimizes ICD sequence configuration by integrating diverse multimodal task augmentation. By systematically enhancing the structure and relevance of ICDs, \textit{SabER} significantly improves ICL performance across multiple LVLMs and VL tasks. Through extensive experiments, we demonstrate that \textit{SabER} outperforms existing SOTA methods and provides new insights into how ICD sequences shape multimodal learning dynamics. Our findings highlight the importance of task-aware sequence configuration and offer a scalable solution to improve the robustness and generalization of multimodal ICL.

\section{Related Works}

\textbf{In-context Learning.} As ICL emerges as an efficient and powerful learning method, research increasingly focuses on its mechanisms \citep{gao2021making, dong2024survey}. \citet{min2022rethink} attribute ICL's success to explicit information in ICDs like label space and input distribution, while \citet{zhou2023least} emphasize the importance of deep input-output mappings for complex tasks. To find a more comprehensive solution, \citet{wei2023larger} and \citet{pan-etal-2023-context} decompose ICL into Task Recognition and Task Learning. \citet{zhao2024unveiling} further propose a two-dimensional coordinate system to explain ICL behavior via two orthogonal variables: similarity in ICDs (perception) and LLMs' ability to recognize tasks (cognition), emphasizing that task-specific semantics in prompt are as crucial as, if not more vital than, sample similarity for effective ICL. 

\textbf{Large Vision-Language Models.} The most representative model with training methods specifically designed for multimodal ICL is the closed-source Flamingo \citep{alayrac2022flamingo}. Its open-source derivative versions, OpenFlamingo \citep{awadalla2023openflamingo} and IDEFICS \citep{laurençon2023obeli}, inherit Flamingo’s strong ICL capabilities and are central to our study. Meanwhile, robust multimodal ICL has become an essential capability of advanced general-purpose LVLMs like InternVL2 \citep{internvl2chen2024far} and Qwen2VL \citep{wang2024qwen2}. To explore and enhance the multimodal ICL of LVLM, recent studies have begun to focus on the interpretability of internal mechanisms, such as research on in-context vectors \citep{huang2024multi, peng2024live}. They inspire further exploration of LVLM workflows and highlight the critical role of ICDs.

\textbf{Configuring ICD sequences.} Due to LLMs' sensitivity to ICD sequences, configuration methods that do not account for the model's ICL mechanisms may degrade overall performance \citep{gao2021making, lu2022fantas}. A notable example is similarity-based retrieval \citep{liu2021makes, sqprli2024configure}. Although this approach has proven effective on certain benchmarks, it underperforms in complex tasks as it fails to provide LLMs with the necessary task-identifying information. Instead, the ICD bias introduced by coarse-grained retrieval amplifies the short-cut effect \citep{lyu2023zicl, yuan2024llm}. Building on these strategies, model-dependent methods have also emerged, employing one or more models for more demanding selection \citep{wu2023self, wang2024learning}. However, these methods often split the retrieval process into multiple steps, lacking an end-to-end approach, thereby increasing complexity. Furthermore, they overly emphasize ICD selection over ordering, highlighting the value of a lightweight autoregressive model for sequence configuration. One work that is closely connected to ours is \citet{yang2024lever}, which introduces a tiny language model composed of two Transformer blocks to automatically select and order ICDs. It neglects the inner mechanisms of reasoning when ICD sequences are input into LVLMs. 

\section{How do LVLMs learn in-context?}
\label{MI}
\subsection{Towards Vision-language ICL}
Following \citep{pan-etal-2023-context} in LLMs, we first attempt to decompose the ICL process of LVLMs into Task Recognition (TR) and Task Learning (TL). In the TR stage, the model uses parametric knowledge to infer the task definition from the ICDs' data distribution. In the TL stage, the model learns the ICDs' content and, with the task semantics from the previous stage, derives the correct input-output mapping. To address the complexity of VL tasks, we aim for a universal ICD representation. Inspired by \citep{si2023measuring}, which shows that semantically specific ICDs can mitigate bias, we design a unified ICD template with a task-relevant intervention, query $Q$. Each ICD can be represented as a triplet $(I, Q, R)$, where $I$ is the image, $R$ is the ground-truth result, and $Q$ is a short task-specific text that instructs models to derive $R$ from $I$. In other words, we explicitly simulate the input-output mapping and add it to the original tuple $(x, y)$. The form and content of $Q$ both vary in different tasks. In this configuration, the query sample is denoted as $(\hat{I},\hat{Q})$.

We develop three settings based on our configuration to examine LVLM's performance on TR or TL separately within open-ended VQA and image classification tasks by manipulating demonstrations: \textbf{Standard, Random, and Dislocation}. (1). \textbf{Standard}: The correct demonstrations $(I_{i},Q_{i},R_{i})$ are used as input to reflect both TR and TL. (2). \textbf{Random}: For a given sequence $S$, all triplets’ $Q$ or $R$ are replaced by the $Q$ or $R$ from one randomly selected demonstration within the sequence. This setting only reflects TR. The two subcategories are represented as Random-Q and Random-R. (3). \textbf{Dislocation}: In this setting, either $Q$ or $R$ in the sequence is modified with content that introduces semantic elements of image captioning task, such as 'describe the whole image,' resulting in $(I_{i}, Q_{i}*, R_{i})$ or $(I_{i}, Q_{i}, R_{i}*)$. This setting only reflects TL. The two subcategories are represented as Dislocation-Q and Dislocation-R. In \textbf{Random} and \textbf{Dislocation}, we specifically avoid altering both $Q$ and $R$, allowing us to compare the individual importance of $Q$ and $R$ to the mechanisms of ICL. We randomly sample $n$-shot demonstrations following a uniform distribution.

\begin{figure}[h]
\centering
  \includegraphics[width=0.4\textwidth]{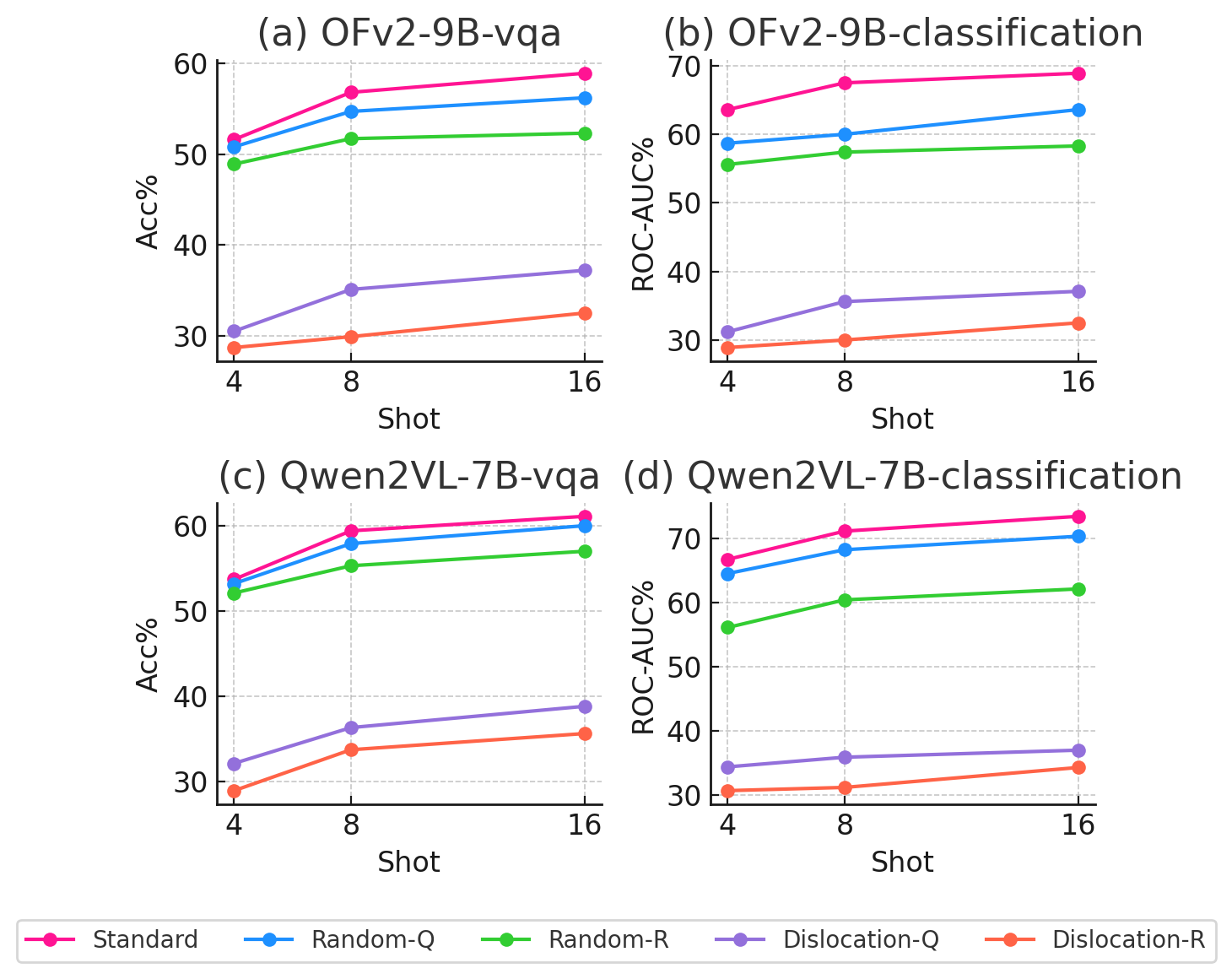}
  \caption{Results of five settings across two tasks and two LVLMs which represent different parts of LVLM's in-context learning.}
  \label{pdis}
\end{figure}

As shown in Figure\ref{pdis}, for both LVLMs, TR is more important than TL because their extensive fine-tuning equips parametric memory to fill TL gaps. However, this may lead to conflicts between internal and external knowledge, emphasizing the need to recognize solid task mapping. TR is more critical for Qwen2VL-7B compared to OFv2-9B , further indicating that differences in the LVLM backbone affect its ability to understand and utilize fine-grained multimodal mapping in TR.

TR is more important than TL because the differences between queries and results are greater, making the mappings within each demonstration and between different demonstrations more difficult to interpret. This implies that the more complex the VL task, the stronger the need for TR, while the demand for TL is prone to be influenced by the LVLM itself. In both TR and TL, $Q$ is more important than $R$, confirming that strong performance in LVLMs is closely related to well-constructed task semantic guidance.

\subsection{Go Deep into TR}
\begin{figure}[h]
    \centering
    \begin{subfigure}{0.35\textwidth}
        \centering
        \raisebox{-5mm}{\includegraphics[width=\linewidth]{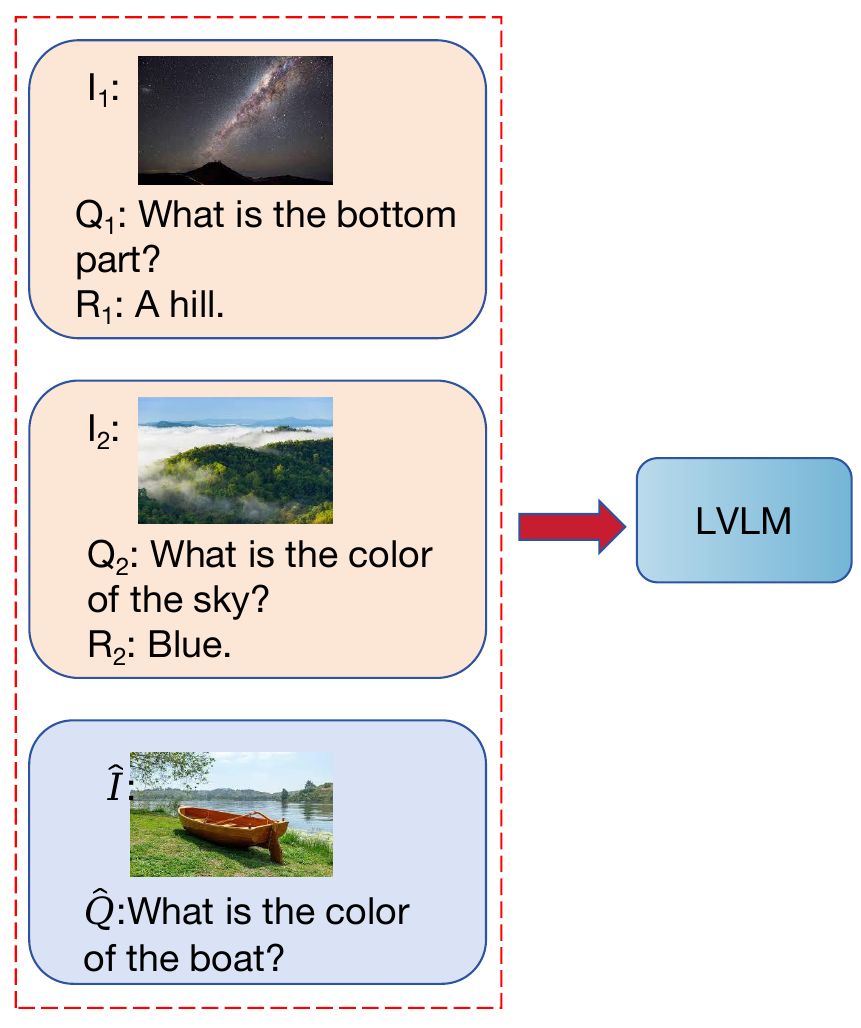}}  
        \caption{2-shot ICD sequence input.}
        \label{fig:input}
    \end{subfigure}
    \hfill
    \begin{subfigure}{0.6\textwidth}
        \centering
        \raisebox{10mm}{\includegraphics[width=\linewidth]{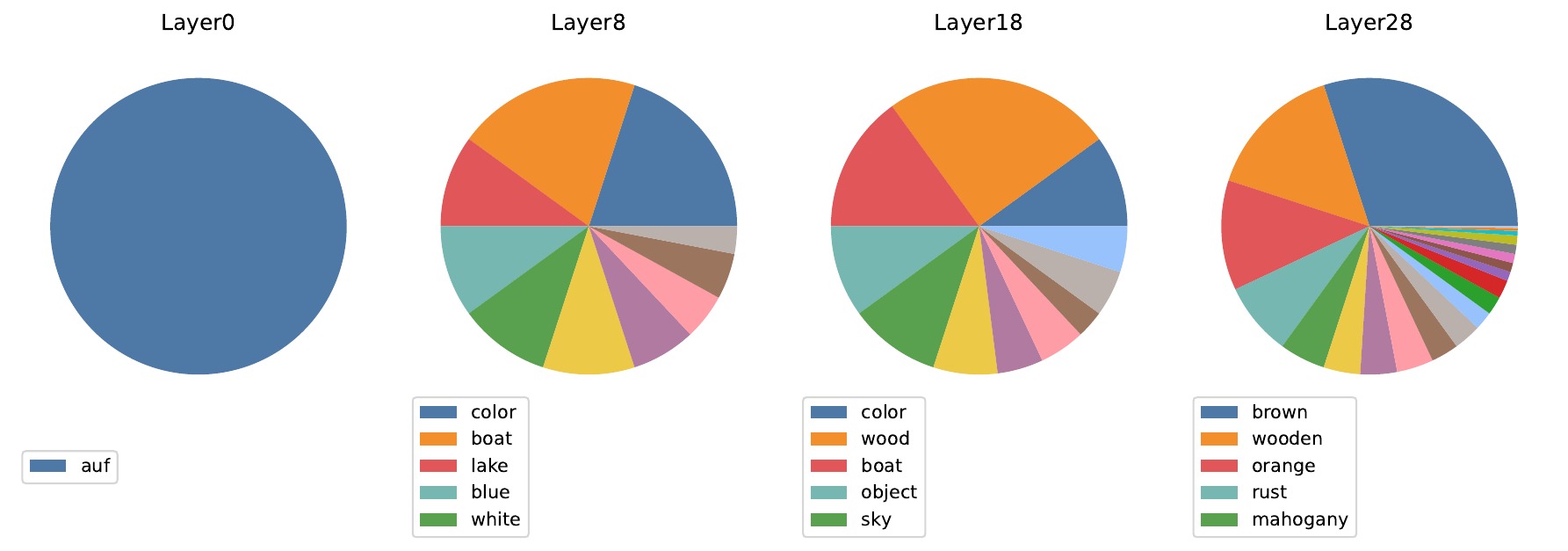}}  
        \caption{Layer-wise output of IDEFICS2.}
        \label{fig:layers}
    \end{subfigure}
    
    \caption{The output of multimodal ICL evolves across layers in the LVLM given a 2-shot sequence (a). As illustrated by the pie charts in (b), processing a complete ICD sequence involves several distinct stages: capturing information from the query sample, identifying mappings within the ICD and engaging in in-depth reasoning, and ultimately leveraging the multimodal information to predict the output.}
    \label{el}
\end{figure}

After identifying the crucial role of TR in multimodal ICL, we further investigate the internal workflow of LVLMs during this stage. Using the logit lens \citep{logit}, we leverage the model’s existing vocabulary space to decode and visualize the last token representation at each layer. Figure \ref{el} illustrates the layer output evolution of IDEFICS2 during multimodal ICL with a given 2-shot ICD sequence. Our findings reveal that TR in multimodal ICL unfolds in two distinct phases: (1). \textbf{Constraining the output space} using the query sample's $\hat{I}$ and $\hat{Q}$, where $Inst$ also plays a role in guiding this process. (2). \textbf{Further refining the output space} by integrating information from all ICDs, including both image and text. Notably, the LVLM does not exhibit a strict order in processing different ICDs within the same sequence. This suggests that all ICDs within a sequence may function collectively. LVLMs do not emphasize cross-modal alignment during TR. However, in the TL stage, alignment information becomes essential, making it essential to ensure proper image-text alignment within each ICD.

\section{Method}
\subsection{Rethinking the Role of ICDs}
\label{rethink}
Based on the analysis in Section \ref{MI}, we conclude that a high-quality ICD sequence maintains a cohesive task mapping that aligns well with the target task mapping of the query sample. This task mapping is collectively formed by all ICDs, meaning that the ICDs function as a unified set, complementing each other rather than being independently stacked in a single direction to create the mapping. The task mapping is further constrained by the instruction and query sample. Thus, a purely similarity-based retrieval approach is insufficient, as it relies solely on embedding-level information, which often introduces inherent limitations. This can result in an ambiguous or even misaligned task mapping, leading to shortcut effects and hallucinations. In conventional ICD sequence configuration, effectively integrating information from existing ICDs, instructions, and query samples simultaneously remains highly challenging. 

To address these challenges, we propose \textit{SabER}, a decoder-only tiny language model that configures ICD sequences with more precise task mapping while maintaining computational efficiency. Using a transformer decoder, \textit{SabER} facilitates the flow of multidimensional task semantics during configuration, ensuring a more coherent and contextually relevant sequence.
\subsection{Model}

\begin{figure*}
    \centering
    \includegraphics[width=0.9\textwidth]{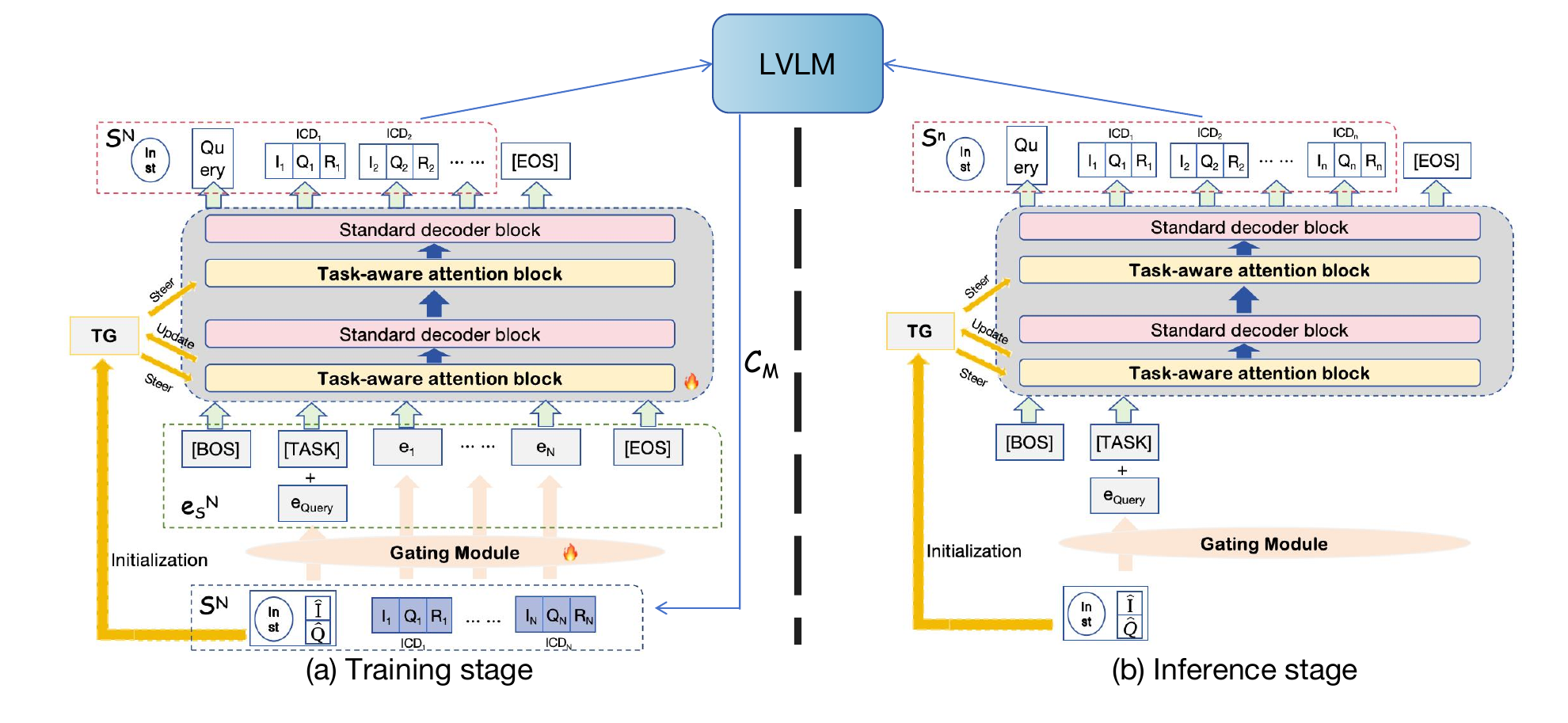}
    \caption{Overview pipeline of our proposed model $SabER$. }
    \label{pipe}
\end{figure*}
Figure \ref{pipe} illustrates the pipeline of \textit{SabER}, which is specifically designed to select ICDs from a demonstration library $DL$ and organize them into sequences in an autoregressive way. \textit{SabER} is centered around four Transformer decoder blocks. Due to its specialized purpose, the vocabulary is entirely composed of samples rather than single words. All tokens correspond one-to-one with each complete sample in $DL$. Consequently, given a query sample as input, \textit{SabER} can progressively retrieve $n$ samples from $DL$ based on the generated token distribution to form the optimal $n$-shot ICD sequence $S^{n}$.

\textbf{Training Data Construction.} 
We construct sequence data for model training using existing high-quality datasets, each corresponding to a VL task (detailed in Section 5). The samples are uniformly formatted as $ (I, Q, R) $ triplets based on their respective task types. Each dataset generates a sequence set $ D_S $ for training, where each sequence consists of a query sample and $ N $ ICDs. The value of $ N $ is configurable, determining the number of shots during training. To ensure optimal training performance, we employ the same LVLM used in inference as a scorer to supervise the construction of $ D_S $, making the method inherently model-specific. For each dataset, we construct $ D_S $ exclusively from its training set through the following three-step process, as detailed in Appendix \ref{mask}.

\textbf{Input Embedding.} 
During training, we aim to clarify the structure of the input sequences in $D_{S}$, composed of ICDs as tokens. To align with the nature of autoregressive generation, we add two special tokens to the vocabulary: $[BOS]$ and $[EOS]$, which represent the beginning and end of a sequence, respectively. We also introduce a $[TASK]$ token into the vocabulary and concatenate it with the query sample in the input sequence. This token enhances the query sample's representation by embedding task-specific information, providing holistic guidance for task recognition. In each \textit{SabER} input sequence, the query sample is positioned ahead of all ICDs. Thus, for a given sequence $S^{N}$, we reconstruct it as $\{[BOS], [TASK] + \hat{x}, x_{1}, ..., x_{N}, [EOS]\}$, which serves as the input sequence to \textit{SabER}. To enable \textit{SabER} to fully obtain essential features from both modality embeddings while maintaining a good balance, we employ a binary gating module to generate the embedding $e_{i}$ for the $i$-th ICD token $x_{i} = (I_{i}, Q_{i}, R_{i})$:
\begin{equation*}
g_{i}=\sigma (W_{g} \cdot [E_{I}(I_{i}) \oplus E_{T}(Q_{i}\oplus R_{i})]+b_{g}),
\end{equation*}
\begin{equation*}
e_{i} = g_{i} \cdot E_{I}(I_{i})+ (1-g_{i}) \cdot E_{T}(Q_{i}\oplus R_{i}),
\end{equation*}
where $E_{I}(\cdot)$ and $E_{T}(\cdot)$ denote image encoder and text encoder of CLIP, respectively. Finally, the input embedding sequence of \textit{SabER} is presented as follows:
\begin{equation*}
e_{S^{N}} = [e_{\text{BOS}}, \hat{e}, e_{1}, \dots, e_{N}, e_{\text{EOS}}],
\end{equation*}
where $e_{\text{BOS}}$ and $e_{\text{EOS}}$ are learnable embeddings defining sequence boundaries. $\hat{e}$ is the joint representation formed by concatenating the learnable embedding of the $[TASK]$ token with the embedding of the query sample $\hat{x}$ generated using the same gating module. In this sequence, the index of $\hat{e}$ is always 1 and $I_{idx}$ denotes the index set of ICD embeddings.

\textbf{Task-aware Attention.} The task-aware attention mechanism in \textit{SabER} enables dynamic configuration of ICD sequences by integrating task semantics into the attention computation. Central to this mechanism is the Task Guider ($TG$), a dedicated embedding that encodes task intent through multimodal fusion of the query sample and instruction:
\begin{equation*}
e_{TG}^{(0)}=W_{TG}\cdot(E_{I}(\hat{I})\oplus E_{T}(\hat{Q}) \oplus E_{T}(Inst')),
\end{equation*}
where $W_{TG} \in \mathbb{R}^{d \times 3d}$ is a learnable weight matrix used to regulate the entire task guider. $Inst'$ is the simplified form of $Inst$ generated by prompting GPT-o1. For clarity, we provide the process of simplification in Appendix \ref{instapp}. This embedding captures the high-level task semantics for the entire sequence.

In predefined task-aware layers $\mathcal{L}_{T}$, $TG$ guides attention through task-semantic relevance weighting. At each layer, $TG$ interacts with token embeddings to compute relevance scores:
\begin{equation*}
    t_i^{(l)} = \sigma\Bigl(\mathrm{MLP}^{(l)}\bigl(e_{TG}^{(l)} \oplus e_i\bigr)\Bigr),
\end{equation*}
where $\mathrm{MLP}^{(l)}$: $\mathbb{R}^{2d} \rightarrow \mathbb{R}^{d}$ is a layer-specific network producing a scalar weight $g_{i}^{l} \in [0,1]$ and $\sigma$ is the sigmoid function. This weight modulates attention logits through a task-aware mask $M^{(l)}$. For intra-ICD tokens, the mask scales pairwise cosine similarities by $log(g_{i}^{(l)})$ to amplify task-critical interactions. Simultaneously, a learnable coefficient $\alpha$ allows the query embedding $\hat{e}$ to steer attention across the entire sequence. Specifically, for position $(i,j)$:
\begin{equation*}
M_{ij}^{(l)} =
\begin{cases}
\frac{\mathrm{sim}(e_i,\, e_j)}{\sqrt{d}} \;\cdot\;
\log\bigl(t_i^{(l)}\bigr),
&  j \le i \text{ and } i,j \in I_{idx}, \\[4pt]
\frac{\alpha \mathrm{sim}(\hat{e},\, e_j)}{\sqrt{d}} \;\cdot\;
\log\bigl(t_1^{(l)} \bigr),
&  i = 1 \text{ and } j \in I_{idx},\\[4pt]
-\infty, 
& \text{otherwise}.
\end{cases}
\end{equation*}

The mask is integrated into standard attention:
\begin{equation*}
\text{Attention}(Q, K, V) = \text{softmax} \left( \frac{QK^T}{\sqrt{d}} + M^{(l)} \right) V.
\end{equation*}

$TG$ is updated only between task-aware layers to preserve task semantic coherence, enable hierarchical refinement from coarse task intent to fine-grained dependencies. After processing layer $l \in \mathcal{L}_{T}$ through residual connections, $TG$ is updated via:
\begin{equation*}
e_{TG}^{(l')} = \operatorname{LN} \left( e_{TG}^{(l)} + \operatorname{Attention} (e_{TG}^{(l)}, H^{(l)}) \right),
\end{equation*}
where $l'$ denotes the next task-aware layer in $\mathcal{L}_{T}$, $H^{(l)}$ denotes the hidden states of layer $l$ and $\operatorname{LN}$ denotes layer normalization. To ensure focused attention patterns, we introduce a sparsity loss that penalizes diffuse attention distributions:
\begin{equation*}
\mathcal{L}_{\text{sparse}} = \sum_{l \in \mathcal{L}_T} \frac{1}{N} \sum_{i=1}^{N} \text{KL} \left( \text{softmax}(M_{i:}^{(l)}) \parallel \mathcal{U} \right),
\end{equation*}
where $\mathcal{U}$ is a uniform distribution. Minimizing this KL divergence forces the model to focus on fewer but semantically salient tokens, enhancing both interpretability and task alignment. The total training objective combines the standard cross-entropy loss for sequence generation, sparsity regularization, and L2-norm constraint on $TG$ to prevent overfitting:
\begin{equation*}
\mathcal{L} = \mathcal{L}_{CE}+\lambda_{1} \mathcal{L}_{\text{sparse}}+\lambda_{2} \left \| W_{TG}\right \| _{2}^{2}.
\end{equation*}

\textbf{Inference and Prompt Construction.} After training \textit{SabER} with $D_{S}$, it can autoregressively select demonstrations from a library and build ICD sequences. Given a query sample $\hat{x} = (\hat{I}, \hat{Q})$, the input sequence to \textit{SabER} during inference is $\{[BOS], [TASK] + \hat{x}\}$, where $\hat{x}$ is embedded using the trained gating module. The number of ICD shots in the generated sequence, denoted as $n$, is a user-defined value. It may differ from the shot count $N$ in $D_{S}$. \textit{SabER} then selects $n$ ICDs, producing the optimal $n$-shot ICD sequence $S^{n}$. This sequence is then used to construct a prompt for LVLMs, formatted as: $(Inst; ICD_{1}, ..., ICD_{n}; Query Sample)$, which is then used to perform multimodal ICL. Example prompts for different LVLMs are provided in Appendix \ref{prompt}.

\section{Experiment}
\begin{table*}
\centering
\resizebox{\textwidth}{!}{\begin{tabular}{ c c c c | c c | c | c | c | c}
\toprule
\multirow{3}*{\textbf{Methods}} & \multicolumn{3}{c|}{\textbf{VQA}} & \multicolumn{2}{c|}{\textbf{Captioning}} & \multicolumn{1}{c|}{\textbf{Classification}} & \multirow{2}*{\textbf{Hybrid}} & \multirow{2}*{\textbf{Fast}} & \multirow{2}*{\textbf{CLEVR}}\\
\cmidrule(lr){2-7}
& VQAv2 & VizWiz & OK-VQA &  Flickr30K & MSCOCO  & HatefulMemes &\multirow{2}*{ACC.↑} & \multirow{2}*{ACC.↑}& \multirow{2}*{ACC.↑}\\
& ACC.↑ & ACC.↑ & ACC.↑ &  CIDEr↑ & CIDEr↑  & ROC-AUC↑ & & &\\
\midrule
  RS & 57.86 & 41.94 & 49.89 & 92.02 & 109.26 & 75.72 & 16.85 & 62.66 & 41.51\\

 I2I & 58.36 & 40.58 & 48.57 & 92.94 & 109.65 & 70.66 & 13.00 & 64.49& 38.63 \\

 IQ2IQ & 59.88 & 43.81 & 51.87 & 93.00 & 109.75& 74.37 & 32.40 &64.47 & 37.37 \\

 IQPR & 59.89 & 42.56 & 51.12 & 94.52& 112.32 & 71.33& 28.67 & 63.99 & 41.00 \\

 Lever-LM & 62.31 & 46.83 & 55.10 & 97.48 & 116.90 & 77.94& 39.29 & 65.02 & 43.66\\
 
 \rowcolor[HTML]{F0F0F0}
 Ours & \textbf{64.74} & \textbf{50.77} & \textbf{57.77}& \textbf{99.42} & \textbf{119.27} & \textbf{79.78} &\textbf{42.93} & \textbf{69.50}& \textbf{46.57}\\
 
\midrule
\% \textbf{Improve} & 3.90\% & 8.41\% & 4.85\% & 2.00\% & 2.03\% & 2.36\% & 9.26\% & 6.89\%& 6.67\%\\
\bottomrule

\end{tabular}}

\caption{\label{main}
Results of different ICD sequence configuration methods across 9 datasets, with both training and generated sequences being 4-shot. Each result is the average performance across five LVLMs with the same prompt format. The highest scores are highlighted in \textbf{bold}. \% \textbf{improve} represents the relative improvement achieved by our model over the previously best baseline. Detailed results for each LVLM can be found in Table \ref{detailed}.
}
\end{table*}
\subsection{Datasets and models}
We first select six high-quality datasets across three key VL tasks and use them as benchmarks to evaluate multimodal ICL: three for open-ended VQA (VQAv2 \citep{vqav2goyal2017making}, VizWiz \citep{vizwizgurari2018vizwiz}, and OK-VQA \citep{okvqamarino2019ok}), two for image captioning (Flickr30K \citep{flickr30kyoung2014image} and MSCOCO \citep{mscocolin2014microsoft}), and one for classification (HatefulMemes \citep{hatefulkiela2020hateful}). For datasets with multiple human-annotated labels per sample, one label is randomly selected as the ground-truth result. To further evaluate \textit{SabER}’s generalization ability of configuring ICD sequences in a complex scenario involving diverse task types, which are more representative of real-world ICL usage contexts \citep{luo2024inde}, we manually create a mixed-task dataset, \textbf{Hybrid}, using the above six datasets. We randomly sample 5,000 instances from each dataset's training set to create \textbf{Hybrid}'s training set, with the validation set proportionally drawn from their validation sets. Towards a more comprehensive evaluation of sequence generation, we also select two challenging image-to-text tasks from the latest multimodal ICL benchmark, VL-ICL \citep{zong2024vlicl}: Fast Open-Ended MiniImageNet (\textbf{Fast}) and \textbf{CLEVR}. See more details in Appendix \ref{dataset}

We experiment with five SOTA LVLMs in total, including four open-source models—Open Flamingo-v2 (9B), IDEFICS2 (8B), InternVL2 (8B), and Qwen2VL (7B)—and one representative closed-source model, GPT-4V \citep{openai2024gpt4v}. These models all support multi-image input and exhibit strong few-shot learning capabilities.

\subsection{Baselines and implementation details}
Given a query sample $\hat{x}=(\hat{I},\hat{Q},\hat{R})$ and a demonstration library $DL$, we compare \textit{SabER} with the following ICD sequence configuration methods: (1). \textbf{Random sampling (RS)}: This method follows a uniform distribution to randomly sample $n$ demonstrations from $DL$. (2). \textbf{Similarity-based retrieval methods}: These methods embed both the demonstrations and the query sample using CLIP, compute cosine similarity under different strategies, and select the top $n$ demonstrations with the highest similarity to construct $S^{n}$. For each demonstration $(I_{i}, Q_{i}, R_{i})$ in $DL$, \textbf{I2I} calculate the similarity solely between $I_{i}$ and $\hat{I}$; \textbf{IQ2IQ} computes the joint similarity between the pairs $(I_{i}, Q_{i})$ and $(\hat{I}, \hat{Q})$; \textbf{IQPR} \citep{sqprli2024configure} evaluates the joint similarity considering all three elements using a pseudo-result $\hat{R}^P$ generated through RS to complete the query sample into a triplet. (3). \textbf{Lever-LM}: A simple tiny language model composed of multiple transformer blocks is trained to perform automatic demonstration selection and construct $S^{n}$. In settings without queries in ICDs, this model outperforms other strategies in VQA and captioning tasks, serving as a key baseline. To ensure a fair comparison, we use a four-layer Lever-LM, which matches the number of layers in \textit{SabER}.

Since we use the training set of each dataset to construct the sequence set $D_{S}$, its validation set is used to evaluate the quality of the ICD sequences generated by \textit{SabER} on the LVLM. We set both the training sequence shot $N$ and the generated sequence shot $n$ at 4. The size of query sample set $K$ varies across different datasets and details can be found in Table \ref{size}. We adopt the image and text encoders from CLIP-ViT-L/14 to generate all image and text embeddings. For all tasks, we adopt a unified encoder training strategy by training only the last three layers while freezing the weights of all preceding layers. During \textit{SabER} training, we apply a cosine annealed warm restart learning scheduler with AdamW as the optimizer, a learning rate set to 1e-4 and a batch size of 128. \textit{SabER} is trained for 20 epochs. 
\begin{table*}
\centering
\resizebox{\textwidth}{!}{\begin{tabular}{lccc|cc|c|c|c|c}
\toprule
\multirow{2}{*}{\textbf{Configuration}} 
& \multicolumn{3}{c|}{\textbf{VQA}} 
& \multicolumn{2}{c|}{\textbf{Captioning}} 
& \textbf{Classification} 
& \multirow{2}{*}{\textbf{Hybrid}}
& \multirow{2}{*}{\textbf{Fast}}
& \multirow{2}{*}{\textbf{CLEVR}} \\
\cmidrule(lr){2-7}
& VQAv2 & VizWiz & OK-VQA
& Flickr30K & MSCOCO
& HatefulMemes 
& &  & \\
\midrule
\textbf{Full \textit{SabER}}       &   \textbf{64.74}    &  \textbf{50.77}     &     \textbf{57.77}  &   \textbf{99.42}    &    \textbf{119.27}   &   \textbf{79.78}    &    \textbf{42.93}   &  \textbf{69.50}     &   \textbf{46.37}    \\
\midrule
(a) w/o [TASK] token   &   62.67    &   48.35    &  55.83     &  97.84   &   117.13    &   77.47    &   39.26    &   67.41    &    44.29   \\
(b) w/o $TG$ updates     &  61.58  &  48.71   &  55.64   &  98.12  &  117.05 &  76.39 &  38.97  &  66.29   &  43.83  \\
(c) w/o Task-aware Mask   &    60.18   & 47.54  & 54.47 &   97.51   &   116.92    &   75.63    &  36.80     &   65.38    &    42.81   \\
\midrule
(d) Random initialization &   55.73    &   37.82    &   47.32    &   93.41    &   105.35    &   71.86   &   29.46   &    59.31   &  40.78    \\
(e) w/o $\hat{I}$  &   61.39    &   47.21    &   54.68    &   96.52    &   114.73    &   76.26    &  37.62     &   66.38    &   43.51    \\
(f) w/o $\hat{Q}$   &   59.46    &  46.07   &   54.05    &   95.78    &   112.61    &   74.32    &  35.87     &    65.49   &   42.35    \\
(g) w/o $Inst'$   &   59.33   &  45.73  &  54.12   &   97.04   &   114.89   &    75.28  &  36.14  & 66.27   &  42.61 \\
\midrule
(h) Layer 1 Only   &   61.78    &    45.26   &   53.97    &   98.35     &   115.82    &   78.10    &  34.45   &   63.49    &    43.17   \\
(i) Layer 3 Only    &   62.67    &   47.52    &   56.38    &   98.84    &   116.68    &   78.72    &  39.63     &    65.57   &    45.04   \\
(j) Layer 2 \& 4   &   63.41    &   48.79    &   56.91    &   99.13    &   118.46    &    78.39   &   41.07    &   67.58    &  45.66   \\
(k) All Layers  &    63.95   &   48.28    &   56.45    &    98.27   &   118.35    &    77.94   &    40.86   &    68.30   &   45.18    \\
\bottomrule
\end{tabular}}
\caption{\label{atable}Results of the ablation study on task augmentation. Each result is the average performance across five LVLMs. Specifically, (a)-(c) correspond to diverse task-aware attention construction, (d)-(g) to diverse $TG$ initialization, and (h)-(k) to diverse placement of task-aware attention.}
\end{table*}

\subsection{Results and Analyses}

Table \ref{main} presents the average ICL performance across five LVLMs with different ICD sequence configuration methods. Notably, \textit{SabER} consistently outperforms all other methods across all nine datasets, showcasing the robustness and efficacy of \textit{SabER} in fully exploiting the potential of LVLMs in multimodal ICL. The performance improvements observed with \textit{SabER} further underline the advantages of augmenting the configuration process with task representations. Specifically, \textit{SabER} yields performance gains ranging from 2.00\% to 9.26\% over the best-performing baselines in various tasks. In VQA, \textit{SabER} delivers an average improvement of 5.72\%, with a notable 8.41\% gain in the challenging VizWiz dataset. The greatest improvement, 9.26\%, is achieved in the mixed-task \textbf{Hybrid} dataset. On \textbf{Fast} and \textbf{CLEVR} designed specifically to benchmark multimodal ICL, \textit{SabER} achieves improvements of 6.89\% and 6.67\%, respectively. These results underscore the importance of leveraging implicit task semantics within ICD sequences, particularly for TR in tasks characterized by diverse or complex mappings. In contrast, simpler tasks, such as image captioning, still benefit from task augmentation, albeit with a more modest average improvement of 2.02\%. We further study the impact of ICD sequence configuration on LVLMs' multimodal ICL performance using the detailed data in Appendix \ref{mainapp}.

\section{Ablation Study}
\subsection{What task-specific augmentation brings?}
In this section, we will focus on the impact of task-aware attention on ICD sequence configuration and its further effect on multimodal ICL in LVLMs.

First, we validate the necessity of task-aware components and hierarchical layer interactions in \textit{SabER}, as shown in Table \ref{atable}. Removing the [TASK] token, which captures task intent, leads to significant performance degradation across question-answering tasks (e.g., VQAv2 drops by 2.07\% and OK-VQA by 2.94\%), as the model struggles to align ICDs with task mapping. Disabling TG updates between layers further degrades performance (e.g., 3.16\% drop on VQAv2), confirming that hierarchical refinement of task semantics is critical for resolving fine-grained dependencies. The task-aware mask, which enforces sparsity in attention patterns, proves indispensable for compositional tasks like HatefulMemes and CLEVR, where its removal causes attention dispersion and reduces accuracy by 6.13\% and 3.56\%, respectively.

Initializing $TG$ with random weights or ablating its multimodal inputs severely undermines task grounding. Random initialization degrades performance catastrophically (VizWiz accuracy drops by 12.95\%), as the model fails to capture task semantics. Query sample's text features seem to be more important than image features $\hat{I}$, though removing both of them results in consistent declines. Instructions semantics is also essential in creating $TG$, and its impact will be further discussed in Section \ref{ablation}. The placement of task-aware attention layers significantly impacts performance. Using only shallow layers (Layer 1) or deep layers (Layer 3) achieves suboptimal results (VQAv2 accuracy: 61.78\% and 62.67\%), as shallow layers lack semantic refinement while deep layers overspecialize. These results collectively emphasize that task-aware agumentation is non-redundant and that their synergistic integration across layers enables robust ICD configuration for diverse vision-language tasks.

To further analyze the impact of task-specific semantics on the entire process, we explore different combinations of training and generation shots, as detailed in Appendix \ref{shot}. 
\subsection{Detailed Analyses}
\label{ablation}
\begin{table}
\centering
\begin{tabular}{lcccccc}
\hline
\textbf{\textit{SabER}} & VQAv2 & MSCOCO & Hatefulmemes & \textbf{Hybrid} & \textbf{Fast} & \textbf{CLEVR} \\
\hline
(CLIP Encoder) &  & &  &  &  &  \\
N/A & 20.41 & 98.26 & 47.82 & 14.80 & 48.67 & 20.52 \\
Adapter only & 25.37 & 108.54 & 67.85 & 18.93 & 54.29 & 25.71 \\
Fully training & \textbf{47.57} & 114.46 & \textbf{76.29}  & \textbf{37.43} & 63.49 & \textbf{43.22} \\
Last two & 42.63 & 114.25 & 73.18 & 28.91 & 62.13 & 39.27 \\
\rowcolor[HTML]{F0F0F0}
Last three & 46.81 & \textbf{114.79} & 75.60 & 35.91 & \textbf{63.72} & 42.18 \\
\hline
(Gating Module) &  & &  &  &  &  \\
+ Ternary gating & 47.21 & 113.92 & \textbf{80.02} & 37.64 & 65.48 & 44.89 \\
\rowcolor[HTML]{F0F0F0}
+ Binary gating & \textbf{50.77} & \textbf{119.27} & 79.78 & \textbf{42.93} & \textbf{69.50} & \textbf{46.57} \\
\hline
\end{tabular}
\caption{\label{emb} Results of \textit{SabER} with different input embedding configurations. (CLIP Encoder) section shows the results without adding gating modules under various training methods for CLIP encoders. N/A indicates no training or modification. (Gating Module) section presents the results with two gating modules added on top of the encoders trained with the method of training the last three layers. The highest scores are highlighted in \textbf{bold}}
\end{table}

\textbf{Input Embedding.} To investigate the impact of input embedding construction on ICD sequence configuration, we vary both the training method of the CLIP encoders and the adoption of the gating module to evaluate \textit{SabER}'s performance under different settings, as detailed in Appendix \ref{embapp}. 

The training approach for CLIP affects the feature representation of embeddings, which in turn influences \textit{SabER}'s ability to capture cross-modal details during sequence configuration. From Table \ref{emb} we observe that for tasks with intrinsic features like VQA and \textbf{Hybrid}, leaving the CLIP unchanged or only adding an adapter leads to significant degradation in the quality of the ICD sequence generation. In fact, even methods that only train the last two layers show a more noticeable performance gap compared to the current approach. This highlights that the output pattern of the third-to-last layer of the encoder is crucial for capturing core task features in multimodal ICD. When we replaced our current training method with one that fully trains CLIP, we did not observe a significant performance drop. This suggests that \textit{SabER}'s treatment of ICDs as tokens does not cause feature loss. In contrast, through task-aware attention, it enhances feature representation, helping mitigate the limitations of the embedding itself. Considering the high cost of training the entire encoders, current method is optimal. As we point out in Section \ref{MI}, it is important for the model to focus on fine-grained features within the two modalities for multimodal ICL. However, Table \ref{emb} shows that the use of a ternary gating mechanism to obtain more refined embeddings actually results in a poorer performance compared to binary gating. 

\textbf{Instruction.} In the ICL workflow of LVLMs, the instruction acts as a general reasoning guide. The results in Table \ref{atable} demonstrate that incorporating the semantics of instructions into $TG$ helps construct a more effective task mapping, resulting in more diverse ICD sequences. However, there is a trade-off between providing detailed instructions and avoiding irrelevant information that may skew task recognition, potentially hindering model convergence. To address this, shortening the instruction using an LLM during $TG$ creation strikes a balance. We test different styles of instruction in Appendix \ref{instapp} and find that the content and format of $Inst$ significantly influence performance, underscoring the importance of its integration into the ICD sequence. Among them, chain-of-thought (CoT) style instructions are the most effective. 
\subsection{Generalization Test}
To showcase the generalization of \textit{SabER} beyond image-to-text tasks, we evaluate its performance on NLP and text-to-image tasks. For NLP tasks, we first use the latest LLM ICL benchmark, ICLEval \citep{chen2024icl}, to organize a mixed-task dataset. This dataset includes all Rule Learning tasks from the benchmark, which are designed to evaluate the ability of LLMs to learn mapping rules from ICDs. We then choose Qwen-7B and LLaMA3-8B as the base LLMs. For text-to-image tasks, we use the Fast Counting dataset from the VL-ICL Bench. We test it on Emu2-Gen \citep{sun2024emu}. The ICDs in both tasks can be represented as $(Q, R)$. In NLP, both $Q$ and $R$ are text; in text-to-image, $Q$ is text while $R$ is an image. We simply need to adjust the embedding encoder and gating module accordingly. The baselines are RS, \textbf{Q2Q} (Query-to-query), \textbf{QPR} (Query\&pseudo-result), and Lever-LM (not applicable to text-to-image). From Table \ref{gen} we observe that \textit{SabER} achieves the best performance across all tasks, proving its excellent generalization and wide application potential.

\begin{table}
\centering
\begin{tabular}{lcc|c}
\hline
\multirow{2}*{\textbf{Methods}} & \multicolumn{2}{c|}{\textbf{NLP}}& \textbf{text-to-image} \\
\cmidrule(lr){2-4}
 & Qwen-7B & LLaMA3-8B & Emu2-Gen\\
\hline
RS & 0.26 & 0.30& 43.67\\
Q2Q & 0.46 & 0.54&  47.83 \\
QPR & 0.45& 0.56 & 49.06 \\
Lever-LM & 0.47 & 0.60 & -\\
\rowcolor[HTML]{F0F0F0}
Ours & \textbf{0.50} & \textbf{0.61} & \textbf{51.18} \\
\hline
\end{tabular}
\caption{\label{gen} Results of different ICD sequence configuration methods in NLP and text-to-image tasks. Both training and generated shots are set to 4. The highest scores are highlighted in \textbf{bold}.}
\end{table}
\section{Conclusion}
We extend LLM research to the multimodal domain, systematically exploring multimodal ICL in LVLMs. We identify a distinct processing logic for interleaved image-text ICDs and emphasize the role of task mapping in sequence configuration. To address this, we propose SabER, a tiny language model that autoregressively selects ICDs and constructs sequences. Guided by theoretical insights, we optimize modality balance and enhance task-semantic interactions with task-aware attention. Extensive experiments validate our approach, demonstrating significant sequence quality improvements and introducing a new perspective on task mapping in multimodal ICL.
\bibliography{iclr2025_conference}
\bibliographystyle{iclr2025_conference}

\appendix

\section{Method}
\subsection{CLIP Encoders}
CLIP employs two distinct encoders: one for images and another for text. The image encoder transforms high-dimensional visual data into a compact, low-dimensional embedding space, using architectures such as a ViT. Meanwhile, the text encoder, built upon a Transformer architecture, generates rich textual representations from natural language inputs.

CLIP is trained to align the embedding spaces of images and text through a contrastive learning objective. Specifically, the model optimizes a contrastive loss that increases the cosine similarity for matched image-text pairs, while reducing it for unmatched pairs within each training batch. To ensure the learning of diverse and transferable visual concepts, the CLIP team curated an extensive dataset comprising 400 million image-text pairs, allowing the model to generalize effectively across various downstream tasks. 

In our experiments, we employ the same model, CLIP-ViT-L/14, using its image and text encoders to generate the image and text embeddings for each demonstration, ensuring consistency in cross-modal representations. The model employs a ViT-L/14 Transformer architecture as the image encoder and a masked self-attention Transformer as the text encoder. We experimented with several strategies for training the CLIP encoder and found that training only the last three layers of the encoder offers the best cost-effectiveness.

\subsection{Demonstration Configuring Details}
\label{demo}
(a) \textbf{Open-ended VQA}: The query $Q_{i}$ is the single question associated with the image $I_{i}$, while the result $R{i}$ is the answer to the question, provided as a short response. For the query sample, $\hat{Q}$ represents the question related to the image $\hat{I}$, and $\hat{R}$ is the expected output of the model. 

(b) \textbf{Image Captioning}: Both $Q_{i}$ and $\hat{Q}$ are set as short prompts instructing the LVLM to generate a caption for the given image, such as "Please write a caption to describe the given image." The result $R_{i}$ corresponds to the actual caption of the image. 

(c) \textbf{Image Classification}: Both $Q_{i}$ and $\hat{Q}$ provide the textual information paired with the image, followed by a directive requiring the model to classify based on the provided image-text pairs. The result $R_{i}$ is the predefined class label. 

For all three tasks mentioned above, since the ground truth answers are not visible to the LVLM during reasoning, all $\hat{R}$ are set to blank.

\subsection{Retrieving Strategies}
Previous works have typically focused on calculating the similarity between either the image or parts of the textual information in the query sample and the demonstrations from the library in isolation. However, this approach can lead to insufficient use of demonstrations by the LVLM, as discussed in Section 3. To address this issue, we propose a fusion-based retrieval strategy \textit{IQ2IQ(image-query to image-query)}, which contains two implementation methods:

(1) \textbf{Averaged Modality Similarity (AMS)} calculate the similarity between
$\hat{I}$ and each $I_{i}$, and between $\hat{Q}$ and each $Q_{i}$, then take the average of these two similarities; 

(2) \textbf{Joint Embedding Similarity (JES)} compute the joint image-text similarity, which concatenates the image and query embeddings to form a comprehensive vector, and use this unified representation to compute the similarity.

\begin{figure*}
    \centering
    \includegraphics[scale=0.4]{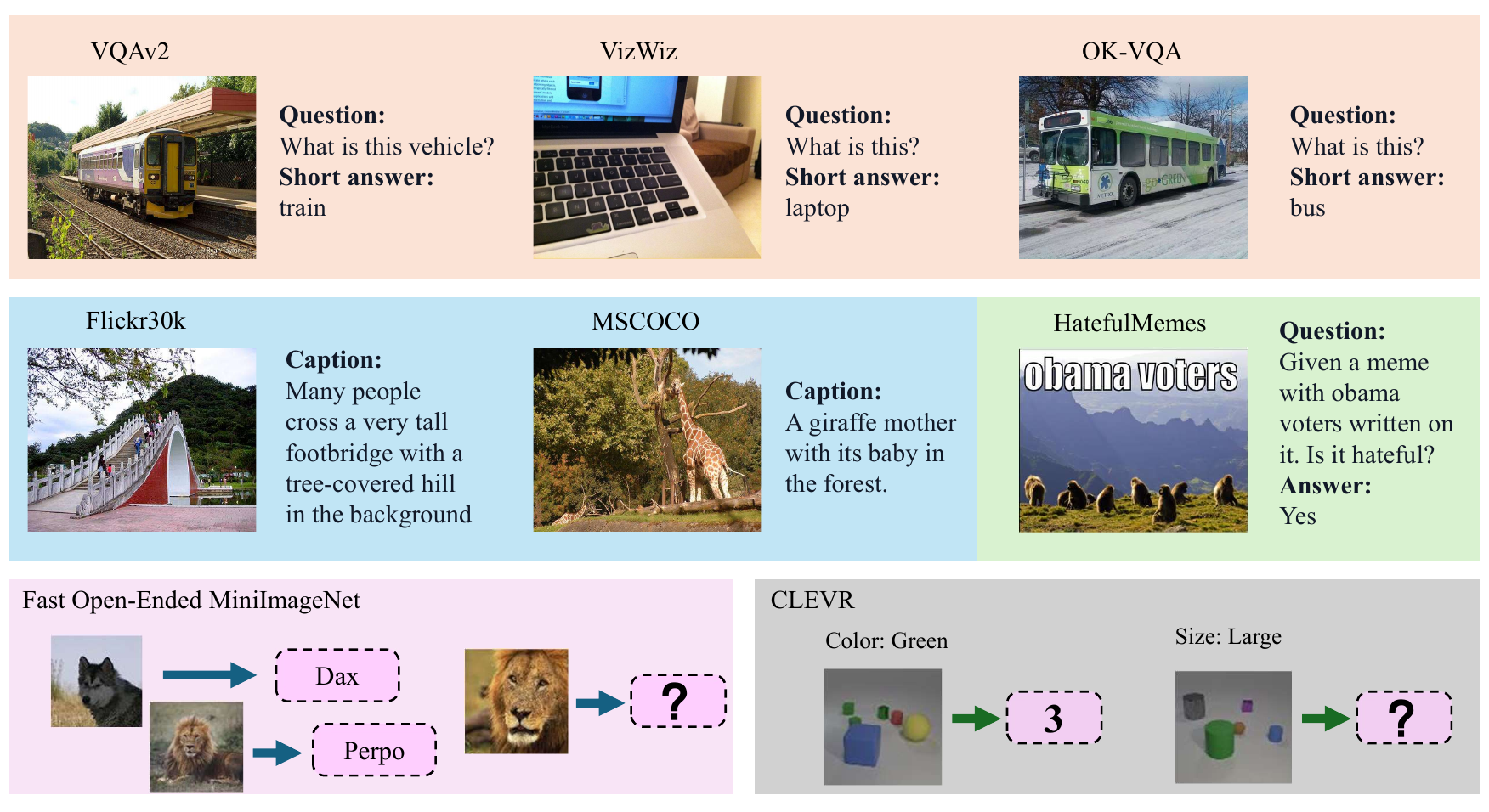}
    \caption{Illustrative examples from various vision-and-language datasets categorized by task type. Visual Question Answering (VQA) tasks are shown in red (VQAv2: train, VizWiz: laptop, OK-VQA: bus). Captioning tasks are represented in blue (Flickr30k: footbridge, MSCOCO: giraffes), while classification tasks are highlighted in green (HatefulMemes: meme identified as hateful). The bottom section demonstrates reasoning tasks with synthetic datasets: Fast Open-Ended MiniImageNet and CLEVR, focusing on conceptual understanding (e.g., assigning labels like "Dax" or identifying object properties like color and size).}
    \label{datasets}
\end{figure*}

\subsection{Instruction}
\label{instapp}
The $Inst$ generated by GPT-4o in the main experiment is "You will be provided with a series of image-text pairs as examples and a question. Your task involves two phases: first, analyze the provided image-text pairs to grasp their context and try to deeply think about what the target task is; second, use this understanding, along with a new image and your knowledge, to accurately answer the given question." This content demonstrates great orderliness and can act as a good general semantic guide for ICDs and query sample. This style is named chain-of-thought (CoT).

To incorporate the semantic information of $Inst$ and strengthen task representation during the ICL sequence configuration process, we use GPT-01 to generate simplified versions of these $Inst$ and integrate their embeddings into the task guider, which are indicated by $Inst'$. The prompt we use is as follows: \textit{"This is an instruction to enable LVLMs to understand and perform a multimodal in-context learning task. Please simplify it by shortening the sentence while preserving its function, core meaning, and structure. The final version should be in its simplest form, where removing any word would change its core meaning".} This simplification process allows us to investigate how the semantic information density in the instruction impacts \textit{SabER}'s sequence configuration ability and the performance of LVLMs in ICL. The results show that simplifying the instruction in a prompt before embedding it in the task guider significantly improves the quality of sequence generation. It also helps to avoid issues caused by too long instructions. 

As shown in Table \ref{instde}, we use GPT-4o to rewrite $Inst$, placing it at the middle and the end of a prompt, altering its semantic structure accordingly while keeping its CoT nature. The table also presents two other tested styles of instructions placed at the beginning of the prompt: Parallel Pattern Integration (PPI) and System-Directive (SD). PPI emphasizes simultaneous processing of pattern recognition and knowledge integration, focusing on dynamic pattern repository construction rather than sequential reasoning. SD structures input as a formal system protocol with defined parameters and execution flows, prioritizing systematic processing over step-by-step analysis. These two forms have also been proven to be effective in previous ICL work. We use them to study the robustness of \textit{SabER} and various LVLMs to different instruction formats.
\begin{table*}

\centering
\begin{tabular}{m{4cm}<{\centering} | m{8cm} }
\toprule
\textbf{$Inst$} & \textbf{Details} \\
\midrule
Beginning1 (CoT) & You will be provided with a series of image-text pairs as examples and a question. Your task involves two phases: first, analyze the provided image-text pairs to grasp their context and try to deeply think about what the target task is; second, use this understanding, along with a new image and your knowledge, to accurately answer the given question. \\

\midrule
Beginning2 (PPI) & Construct a dynamic pattern repository from image-text samples, then leverage this framework alongside your knowledge base for concurrent visual analysis and question resolution. The key is parallel processing - your pattern matching and knowledge integration should happen simultaneously rather than sequentially. \\

\midrule
Beginning3 (SD)& SYSTEM DIRECTIVE
Input Stream: Example Pairs → New Image + Query
Process: Pattern Extract → Knowledge Merge → Visual Analysis → Response Critical: All exemplar patterns must inform final analysis Priority: Context preservation essential\\

\midrule
Middle & Now you have seen several examples of image-text pairs. Next, you will be given a question. Your task involves two phases: first, revisit the above image-text pairs and try to deeply think about what the target task is; second, use this understanding, along with a new image and your knowledge, to accurately answer the given question. \\

\midrule
End & Now you have seen several examples of image-text pairs and a question accompanied by a new image. Your task involves two phases: first, revisit the provided examples and try to deeply think about what the target task is; second, use this understanding, the new image and your knowledge to accurately answer the given question. \\

\midrule
Beginning1 (Abbreviated) & Analyze the following image-text pairs, understand the task, and use this to answer the question with a new image. \\
\midrule
Middle (Abbreviated) & After reviewing the above image-text pairs, analyze the task and use this understanding to answer the question with a new image. \\
\midrule
End (Abbreviated) & After reviewing the above image-text pairs and a question with a new image, analyze the task and use this understanding it. \\

\bottomrule

\end{tabular}
\caption{\label{instde} Formats of different instruction types and their corresponding details used in the prompt structure for all VL tasks. (Abbreviated) means that the instruction is a simplified version produced by GPT-o1. }

\end{table*}

\subsection{Prompt Details}
\label{prompt}
The prompts constructed based on $S^{n}$ all follow the format:
$$(Inst; ICD_{1}, ..., ICD_{n}; Query Sample).$$ 
Each ICD's query begins with "Question:" and its result starts with "Answer:". The query sample concludes with "Answer:", prompting the LVLM to generate a response. Depending on the input format required by different LVLMs, we may also include special tags at the beginning and end of the prompt. 

Table \ref{prompt-details} provides an overview of the prompt details used for the different models in our experiments. Each model, including OpenFlamingov2, ICDEFICSv2, InternVL2, and Qwen2VL, employs a structured approach to engage with image-text pairs. The two-phase task requires LVLMs to first absorb information from a series of prompts before utilizing that context to answer subsequent questions related to new images. This method allows for enhanced understanding and reasoning based on prior knowledge and context, which is essential for accurate question answering in vision-and-language tasks.
\begin{table*}
\centering
\begin{tabular}{m{3cm}<{\centering} | m{9cm}}
\toprule
\textbf{Models} & \textbf{Prompt details} \\
\midrule
OpenFlamingov2 & Your task involves two phases: first, analyze the provided image-text pairs to grasp their context and try to deeply think about what the target task is; second, use this understanding, along with a new image and your knowledge, to accurately answer an upcoming question.

$$<$$img>$<$IMG\_CONTEXT>$<$|endofchunk|>
Question: In what country can you see this?
Answer: vietnam

$<$img>$<$IMG\_CONTEXT>$<$|endofchunk|>
Question: Is this a buggy or car?
Answer: buggy

$<$img>$<$IMG\_CONTEXT>$<$|endofchunk|>
Question: What is this?
Answer: \\
\midrule
IDEFICS2 & "User: Your task involves two phases: first, analyze the provided image-text pairs to grasp their context and try to deeply think about what the target task is; second, use this understanding, along with a new image and your knowledge, to accurately answer an upcoming question."

"\textbackslash{}nUser:$<$|image\_pad|> Question: In what country can you see this? $<$end\_of\_utterance>",

"\textbackslash{}nAssistant: Answer: vietnam. $<$end\_of\_utterance>",

"\textbackslash{}nUser: $<$|image\_pad|> Question: Is this a buggy or car? $<$end\_of\_utterance>",

"\textbackslash{}nAssistant:  Answer: buggy. $<$end\_of\_utterance>",

$<$|image\_pad|> Question: What is this? $<$end\_of\_utterance>",

"\textbackslash{}nAssistant: Answer:" \\
\midrule
InternVL2 & Your task involves two phases: first, analyze the provided image-text pairs to grasp their context; second, use this understanding, along with a new image and your knowledge, to accurately answer an upcoming question.

$<$img>$<$IMG\_CONTEXT>$<$/img>
Question: In what country can you see this?
Answer: vietnam

$<$img>$<$IMG\_CONTEXT>$<$/img>
Question: Is this a buggy or car?
Answer: buggy

$<$img>$<$IMG\_CONTEXT>$<$/img>
Question: What is this?
Answer: \\
\midrule
Qwen2VL & $<$|im\_start|>system

You are a helpful assistant.$<$|im\_end|>

$<$|im\_start|>user

Your task involves two phases: first, analyze the provided image-text pairs to grasp their context and try to deeply think about what the target task is; second, use this understanding, along with a new image and your knowledge, to accurately answer an upcoming question.

$<$|vision\_start|>$<$|image\_pad|>$<$|vision\_end|>Question:In what country can you see this? Answer: vietnam

$<$|vision\_start|>$<$|image\_pad|>$<$|vision\_end|>Question: Is this a buggy or car? Answer: buggy

$<$|vision\_start|>$<$|image\_pad|>$<$|vision\_end|>Question: What is this? Answer: $<$|im\_end|>

$<$|im\_start|>assistant \\

\bottomrule

\end{tabular}
\caption{\label{prompt-details} Prompt details for different models used in the experiments. The table outlines how OpenFlamingov2, IDEFICS2, InternVL2, and Qwen2-VL format their image-text interactions, including examples of image-based questions and short answers. Each model follows a multi-phase task structure, where context is absorbed from previous image-text pairs to answer subsequent questions.}

\end{table*}

\subsection{Model}
\label{mask}
\textbf{Training Data Construction.} (1). We apply $ k $-means clustering based on image features to partition the dataset into $ k $ clusters. From each cluster, we select the $ m $ samples closest to the centroid, yielding a total of $ K = m \times k $ samples. These form the query sample set $ \hat{D} $ after removing their ground-truth results, which are stored separately in $ D_{\hat{R}} $. The remaining dataset serves as the demonstration library $ DL $. (2). For each query sample $ \hat{x}_i \in \hat{D} $, we randomly sample a candidate set $ D_i $ of $ 64n $ demonstrations from $ DL $. The objective is to retrieve $ N $ demonstrations from $ D_i $ that optimally configure the sequence for $ \hat{x}_i = (\hat{I}_i, \hat{Q}_i) $ with its ground-truth result $ \hat{R}_i = (\hat{R}_i^{(1)},..., \hat{R}i^{(t)}) $. We use the log-likelihood score computed by the LVLM $ \mathcal{M} $ as the selection criterion $ \mathcal{C}_\mathcal{M} $, evaluating the model's predictive ability given a sequence with $ n $ ICDs:
\begin{equation*}
\mathcal{C}_\mathcal{M}(S^{n}_{i})=\sum_{t}^{} log P_{\mathcal{M}}(\hat{R}_{i}^{(t)} \mid S^{n}_{i},\hat{R}_{i}^{(1:t-1)}),
\end{equation*}

To determine the optimal $ n $-th demonstration $ x_n $ for a sequence $ S_i^{n-1} $ with $ n-1 $ ICDs, we select the candidate that maximizes the incremental gain in $ \mathcal{C}_\mathcal{M} $:

\begin{equation*}
x_{n}= \underset {x\in D_{i}}{argmax}[\mathcal{C}_\mathcal{M}(S^{n-1}_{i}+x)-\mathcal{C}_\mathcal{M}(S^{n-1}_{i})].
\end{equation*}

(3). We employ beam search with a beam size of $ 2N $, ensuring that for each $ \hat{x} $, the top $ 2N $ optimal sequences are included in $ D_S $. As a result, the final sequence set $ D_S $ consists of $ 2N \times k $ $ N $-shot sequences, providing refined training data for the model.

\section{Experiment}
\label{dL}
\subsection{Dataset}
\label{dataset}
\begin{table*}

\centering
\resizebox{\textwidth}{!}{\begin{tabular}{c | c c c c c c c c c}
\toprule
\textbf{Datasets} & \textbf{VQAv2} & \textbf{VizWiz} & \textbf{OK-VQA} & \textbf{Flickr30k} & \textbf{MSCOCO} & \textbf{HatefulMemes} & \textbf{Hybrid} & \textbf{Fast} & \textbf{CLEVR}\\
\midrule
metrics & Accuracy & Accuracy & Accuracy & CIDEr & CIDEr & ROC-AUC & Accuracy & Accuracy & Accuracy\\
\bottomrule
\end{tabular}}
\caption{\label{metrics} Evaluation metrics used for each dataset. Accuracy is used for VQA datasets (VQAv2, VizWiz, OK-VQA), self-bulit \textbf{Hybrid} dataset and two VL-ICL Bench's tasks. CIDEr \citep{cidervedantam2015cider} is used for image captioning datasets (Flickr30k, MSCOCO). ROC-AUC is used for the HatefulMemes classification task.}

\end{table*}
In our study, we explore various VL tasks that use diverse datasets to evaluate model performance. As illustrated in Figure \ref{datasets}, we use VQA datasets such as VQAv2, VizWiz, and OK-VQA, which test the models' abilities in question-answer scenarios. Additionally, we incorporate image captioning datasets such as Flickr30k and MSCOCO to assess descriptive accuracy, along with the HatefulMemes dataset for classification tasks focused on hate speech detection. This comprehensive approach allows us to thoroughly evaluate the models across different tasks. The size distribution of the training, validation, test and query sets $\hat{D}$ in these VL datasets is shown in Table \ref{size}.

\begin{table}
\centering
\begin{tabular}{c c c c c}
\toprule
\textbf{Datasets} & \textbf{Training} & \textbf{Validation} & \textbf{Test} & \textbf{$\hat{D}$ Size}\\
\midrule
VQAv2 & 443,757 & 214,354 & 447,793& 8000\\
VizWiz & 20,523 & 4,319 & 8,000&2000\\
OK-VQA & 9,055 & 5,000 & /&800\\
Flickr30k & 29,783 & 1,000 & 1,000&2500\\
MSCOCO & 82,783 & 40,504 & 40,775&3000\\
HatefulMemes & 8,500 & 500 & 2,000&800\\
\textbf{Hybrid} & 30000 & 9000 & /& 3000\\
\textbf{Fast} & 5,000 & / & 200& 500\\
\textbf{CLEVR} & 800 & / & 200& 80\\
\bottomrule
\end{tabular}
\caption{\label{size} Overview of the size distribution across the datasets used.}
\end{table}

For the Open-ended VQA task, we utilize the following datasets: VQAv2, which contains images from the MSCOCO dataset and focuses on traditional question-answering pairs, testing the model's ability to understand both the image and the question. VizWiz presents a more challenging setting with lower-quality images and questions along with a lot of unanswerable questions, pushing models to handle uncertainty and ambiguity. OK-VQA is distinct in that it requires the model to leverage external knowledge beyond the image content itself to generate correct answers, making it a benchmark for evaluating models’ capacity to integrate outside information.

For the Image Captioning task, we use the Flickr30k and MSCOCO datasets. The Flickr30k dataset consists of images depicting everyday activities, with accompanying captions that provide concise descriptions of these scenes. The MSCOCO dataset is a widely-used benchmark featuring a diverse range of images with detailed and richly descriptive captions, ideal for evaluating image captioning models.

For the Image Classification task, we use the HatefulMemes dataset, which is an innovative dataset designed to reflect real-world challenges found in internet memes. It combines both visual and textual elements, requiring the model to jointly interpret the image and the overlaid text to detect instances of hate speech.

VL-ICL Bench covers a number of tasks, which includes diverse multimodal ICL capabilities spanning concept binding, reasoning or fine-grained perception. Few-shot ICL is performed by sampling the ICDs from the training split and the query examples from the test split. We choose two image-to-text generation tasks from it, which reflects different key points of ICL. Fast Open MiniImageNet task assigns novel synthetic names (e.g., dax or perpo) to object categories, and LVLMs must learn these associations to name test images based on a few examples instead of their parametric knowledge, emphasizing the importance of rapid learning from ICDs. CLEVR Count Induction asks LVLMs to solve tasks like \textit{"How many red objects are there in the scene?"} from examples rather than explicit prompts. The ICDs' images are accompanied by obscure queries formed as attribute-value pairs that identify a specific object type based on four attributes: size, shape, color, or material. Models must perform challenging reasoning to discern the task mapping and generate the correct count of objects that match the query attribute.

The datasets in our experiments are evaluated using task-specific metrics, as summarized in Table \ref{metrics}. For the VQA tasks, \textbf{Hybrid} dataset and VL-ICL bench's tasks, we use accuracy as the metric to assess the models' ability to provide correct answers:
\begin{equation*}
Acc_{a_{i}}=max(1,\frac{3 \times  {\textstyle \sum_{k\in [0,9]}^{}}match(a_{i},g_{k}) }{10} ),
\end{equation*}
where $a_{i}$ denotes the model's generated answer, $g_{k}$ denotes the $k$-th ground true answer. $match(\cdot,\cdot)$ decides whether two answers match, if they match, the result is 1, otherwise it is 0.

For the image captioning tasks, we use the CIDEr score, which measures the similarity between generated captions and human annotations. Finally, for the HatefulMemes classification task, we evaluate performance using the ROC-AUC metric, which reflects the model's ability to distinguish between hateful and non-hateful content.

\subsection{LVLMs}
\label{LVLM}
In recent advances of large vision language models (LVLMs), efficient processing of multimodal inputs, especially images, has become a critical focus. Models like OpenFlamingov2, IDEFICS2, InternVL2, Qwen2-VL and GPT-4V implement unique strategies to manage and process visual data alongside textual input.

OpenFlamingov2 handles visual input by dividing images into patches and encoding them with a Vision Transformer. Each image patch generates a number of visual tokens, which are then processed alongside text inputs for multimodal tasks. To manage multi-image inputs, the model inserts special tokens $<$image> and $<$|endofchunk|> at the beginning and end of the visual token sequences. For example, an image divided into 4 patches produces 4 x 256 visual tokens, with the additional special tokens marking the boundaries before the tokens are processed by the large language model. 

IDEFICS2 processes visual input by applying an adaptive patch division strategy adapted to image resolution and content complexity. Depending on these factors, each image is segmented into 1 to 6 patches, striking a balance between preserving spatial information and maintaining efficiency. These patches are encoded through a Vision Transformer, followed by a spatial attention mechanism and a compact MLP, resulting in 128 visual tokens per patch. The positions of images in the input sequence are marked with $<$|image\_pad|> for alignment, while $<$end\_of\_utterance> tokens separate query and answer components in in-context demonstrations. An image split into five patches yields 5 x 128 + 2 tokens before being integrated with the LLM. 

InternVL2 also dynamically divides images into 1 to 4 patches based on their aspect ratio. A Vision Transformer then extracts visual features from each patch, followed by a pixel shuffle operation and a mlp, producing 256 visual tokens for each patch. Additionally, special tokens $<$img> and $<$/img> are inserted at the beginning and end of the sequence. So, an image divided into 3 patches will produce 3 x 256 + 2 tokens before entering LLM.

Qwen2-VL reduces the number of visual tokens per image through a compression mechanism that condenses adjacent tokens. A ViT first encodes an image (e.g., with a resolution of 224 x 224 and a patch size of 14), producing a grid of tokens, which is then reduced by employing a simple MLP to compress 2 x 2 tokens into a single token. Special $<$lvision\_start|> and $<$lvision\_end|> tokens are inserted at the start and end of the compressed visual token sequence. For example, an image that initially generates 256 visual tokens is compressed to just 66 tokens before entering the LLM. 

GPT-4V (Vision) extends GPT-4's capabilities to handle VL tasks by enabling the model to process and reason about visual input alongside text. The model can perform various tasks including image understanding, object recognition, text extraction, and visual question-answering through natural language interaction. In terms of its few-shot learning ability, GPT-4V demonstrates the capacity to adapt to new visual tasks given a small number of examples through natural language instructions, showing potential in areas such as image classification and visual reasoning, though performance may vary across different task domains and complexity levels.

\subsection{Baseline}
\label{baseline}
Various baseline methods are used to evaluate the model's performance, ranging from random sample to different SOTA retrieval strategies. The following is a description of the baselines used in our experiments.

1. \textbf{Random Sampling (\textbf{RS})}: In this approach, a uniform distribution is followed to randomly sample $n$ demonstrations from the library. These demonstrations are then directly inserted into the prompt to guide the model in answering the query.

2. \textbf{Image2Image (I2I)}: During the retrieval process, only the image embeddings $I_{i}$ from each demonstration $(I_{i}, Q_{i}, R_{i}$ are used. These embeddings are compared to the query image embedding $\hat{I}$ and the retrieval is based on the similarity between the images.

3. \textbf{ImageQuery2ImageQuery (IQ2IQ)}: During the retrieval process, both the image embeddings $I_{i}$ and the query embeddings $Q_{i}$ of each demonstration $(I_{i}, Q_{i}, R_{i}$ are used. These embeddings are compared to the embedding of the concatenated query sample $(\hat{I},\hat{Q})$ and the retrieval is based on the joint similarity between the images and the queries.

4. \textbf{ImageQuery\&Pseudo Result (IQPR)}: This baseline starts by using the RS to generate a pseudo result $\hat{R}^{P}$ of the query sample. The pseudo result is then concatenated with $\hat{I}$ and $\hat{Q}$ to form the query sample's embedding. This retrieval method is based on the similarity of the whole triplet, using image, query and result embeddings.

5. \textbf{Lever-LM}: Lever-LM is designed to capture statistical patterns between ICDs for an effective ICD sequence configuration. Observing that configuring an ICD sequence resembles composing a sentence, Lever-LM leverages a temporal learning approach to identify these patterns. A special dataset of effective ICD sequences is constructed to train Lever-LM. Once trained, its performance is validated by comparing it with similarity-based retrieval methods, demonstrating its ability to capture inter-ICD patterns and enhance ICD sequence configuration for LVLMs.

\subsection{Main Results}
\label{mainapp}
\begin{table*}
\centering
\resizebox{\columnwidth}{!}{\begin{tabular}{c | c c c c | c c | c | c | c | c}
\toprule
\multicolumn{2}{c}{} & \multicolumn{3}{c|}{\textbf{VQA}} & \multicolumn{2}{c|}{\textbf{Captioning}} & \multicolumn{1}{c|}{\textbf{Classification}} & \multirow{2}*{\textbf{Hybrid}} & \multirow{2}*{\textbf{Fast}} & \multirow{2}*{\textbf{CLEVR}}\\
\cmidrule(lr){3-8}
\multicolumn{2}{c}{} & VQAv2 & VizWiz & OK-VQA &  Flickr30K & MSCOCO  & HatefulMemes &~ & ~& ~\\
\midrule
\multirow{6}*{OpenFlamingov2} & RS & 49.52 & 27.71 & 37.90 & 76.74 & 92.98 & 70.53 & 13.48 & 57.69 & 21.60 \\

~ & I2I & 50.84 & 26.82 & 37.79 & 79.84 & 94.31 & 64.75 & 12.79 & 59.07 & 19.39 \\

~ & IQ2IQ & 52.29 & 31.78 & 42.93 & 79.91 & 94.40 & 68.72 & 24.93 & 58.96 & 20.03\\

~ & SQPR & 53.38 & 30.12 & 41.70 & 80.02 & 96.37 &  69.16 & 28.71 & 57.32 & 21.84 \\

~ & Lever-LM & 55.89 & 33.34 & 43.65 & 83.17 & 98.74 & 72.70 & 32.04 & 59.41 & 22.67\\

~ & Ours & \textbf{60.12} & \textbf{39.76} & \textbf{46.28} & \textbf{84.23} & \textbf{99.10} & \textbf{75.09} & \textbf{35.17}& \textbf{62.25} & \textbf{26.80} \\
\midrule
\multirow{6}*{IDEFICS2} & RS & 53.77 & 32.92 & 40.01 & 82.43 & 99.61 & 68.81 & 15.65 & 54.72 & 35.14 \\

~ & I2I & 54.97 & 31.67 & 41.37 & 85.76 & 101.34 & 69.31 & 10.49 & 55.20 & 32.37 \\

~ & IQ2IQ & 55.41 & 34.31 & 43.13 & 85.63 & 101.45 & 70.78 & 30.36 & 55.14 & 32.75\\

~ & SQPR & 55.32 & 33.74 & 42.76 & 87.65 & 103.57 & 62.18 & 24.03 & 55.18 & 36.29\\

~ & Lever-LM & 56.78 & 34.10 & 43.27 & 88.01 & 105.62 & 71.33 & 30.14 & 55.83 & 38.97 \\

~ & Ours & \textbf{58.41} & \textbf{38.32} & \textbf{47.35} & \textbf{90.41} & \textbf{107.04} & \textbf{73.68} & \textbf{33.25}&\textbf{61.21} & \textbf{40.21} \\
\midrule
\multirow{6}*{InternVL2} & RS & 61.83 & 54.70 & 57.13 & 99.05 & 116.37 & 76.84 & 17.74 & 75.87 & 57.03 \\

~ & I2I & 63.35 & 55.07 & 58.73 & 103.29  & 118.46 & 70.72 & 14.82 & 75.89 & 54.79 \\

~ & IQ2IQ & 64.57  & 56.94 & \textbf{62.91} & 103.41& 118.53 & 78.20 & 36.46 & 76.03 & 50.07 \\

~ & SQPR & 63.67 & 56.83 & 60.14 & 105.28 & 121.94 &  77.31 & 34.05 & 76.34 & 56.32\\

~ & Lever-LM & 65.36 & 57.27 & 61.11 & 104.65 & 126.12 & 79.58 & 43.16 & 78.84 & 57.45 \\

~ & Ours & \textbf{68.42} & \textbf{61.69} & 62.87& \textbf{108.26} & \textbf{128.34} & \textbf{82.97} & \textbf{45.79} &\textbf{81.76}  & \textbf{59.27} \\
\midrule
\multirow{6}*{Qwen2VL} & RS & 63.71 & 48.97& 55.30 & 100.32 & 121.47 & 80.01 & 20.42& 66.29 & 48.70\\

~ & I2I & 64.28 & 48.75 & 56.39 & 102.87 & 124.50 & 77.85 & 13.89 & 67.81 & 47.97\\

~ & IQ2IQ & 67.26 & 52.20 & 58.49 & 103.04 & 124.63 & 79.78 & 37.83 & 67.76 & 46.63\\

~ & SQPR & 67.49 & 49.54 & 59.86 & 105.13 & 127.38& 76.67 & 27.96 & 67.12 & 49.56\\

~ & Lever-LM & 68.23 & 54.81 & 61.75 & 105.24 & 127.03 & 81.29 & 45.47 & 70.73 & 50.85 \\

~ & Ours & \textbf{71.57} & \textbf{57.93} & \textbf{63.97} & \textbf{106.91} & \textbf{132.14} & \textbf{83.19} & \textbf{48.95} & \textbf{75.09} &\textbf{55.98} \\

\midrule
\multirow{6}*{GPT-4V} & RS & 60.49 & 45.38 & 59.13 & 101.56 & 115.87 & 82.40 & 16.98 & 58.72 & 45.08\\

~ & I2I & - & - &- & - & - & - & -& -& - \\

~ & IQ2IQ & - & - &- & - & - & - & -& -& -  \\

~ & SQPR & - & - &- & - & - & - & -& -& - \\

~ & Lever-LM & \textbf{65.31} & 54.62 & 65.73 & 106.34 & 126.98 & \textbf{84.81} & 45.62 & 60.31& 48.34 \\

~ & Ours & 65.16 & \textbf{56.17} & \textbf{68.39} & \textbf{107.29} & \textbf{129.71} & 83.96 &\textbf{51.48} & \textbf{67.17} & \textbf{50.59}\\

\bottomrule

\end{tabular}}
\caption{\label{detailed}
Detailed results of different methods across all tasks for the five LVLMs used in the evaluation, with all generated sequences being 4-shot. The highest scores are highlighted in \textbf{bold}. Our model achieves the best performance in all but three tasks, demonstrating its generalization and effectiveness.
}
\end{table*}
We can go deep into the results in Tabel \ref{detailed}. The findings are as follows: (1) \textit{SabER} exhibits the best performance in all but three tasks across nine datasets and five LVLMs, demonstrating its great efficiency and generalization. Upon examining the outputs, we observe that GPT-4V tends to deviate from the ICD format and produce redundant information more easily than open-source LVLMs, aligning with \citep{wu2023gpt4v}. This results in the quality improvement of the ICD sequence not always translating into stable ICL performance gains for GPT-4V, which may explain why \textit{SabER} did not achieve the best performance in two of its tasks. (2) For tasks like VizWiz and \textbf{Hybrid}, \textit{SabER} consistently improves the quality of sequence generation in all LVLMs compared to similarity-based models, demonstrating the importance of increasing task semantics for complex task mappings. We find that the performance gains from \textit{SabER} are not directly related to the model's intrinsic ability on these tasks. Unlike simpler tasks like captioning, for tasks with complex mappings, task semantics still has a significant impact, even when LVLMs exhibit strong few-shot learning abilities. This shows that models with strong ICL capabilities on certain tasks retain, and even strengthen, their ability to leverage task semantics, underscoring the value of improving ICD sequence quality.
\section{Ablation Study}
\subsection{Devil in Shot Counts}
\begin{table*}
\centering
\resizebox{\textwidth}{!}{\begin{tabular}{cccc|ccc|ccc|ccc|ccc|ccc}
\toprule
 & \multicolumn{3}{c}{VizWiz} & \multicolumn{3}{c}{MSCOCO} & \multicolumn{3}{c}{Hatefulmemes} & \multicolumn{3}{c}{\textbf{Hybrid}} & \multicolumn{3}{c}{\textbf{FAST}} & \multicolumn{3}{c}{\textbf{CLEVR}} \\
\cmidrule(lr){2-4} \cmidrule(lr){5-7} \cmidrule(lr){8-10} \cmidrule(lr){11-13} \cmidrule(lr){14-16} \cmidrule(lr){17-19}
 \diagbox{N}{n} & 2 & 4 & 8 & 2 & 4 & 8 & 2 & 4 & 8 & 2 & 4 & 8 & 2 & 4 & 8 & 2 & 4 & 8 \\
\midrule
\multirow{2}{*}{2} & 50.25 & 50.91 & 50.68 & 115.56 & 121.05 & 120.72 & 78.77 & 82.93 & 81.52 & 37.27 & 43.15 & 42.68 & 68.15 & 71.33 & 73.07 & 44.16 & 47.58 & 48.93 \\
 & (13.16\%) & (10.74\%) & (11.66\%) & (6.76\%) & (5.44\%) & (5.46\%) & (5.07\%) & (4.12\%) & (3.88\%) & (11.73\%) & (9.24\%) & (9.77\%) & (7.91\%) & (7.69\%) & (6.66\%) & (8.38\%) & (7.73\%) & (7.19\%) \\
\midrule
\multirow{2}*{4} & 49.69 & \underline{54.17} & 55.83 & 117.79 & \underline{122.67} & 114.21 & 77.64 & \underline{83.18} & 84.89 & 34.10 & \underline{46.33} & 47.05 & 69.88 & \underline{72.90} & 73.63 & 42.94 & \underline{49.97} & 49.79 \\
 & (11.99\%) & (\underline{13.63\%}) & (12.55\%) & (4.37\%) & (\underline{4.77\%}) & (4.68\%) & (3.52\%) & (\underline{5.21\%}) & (5.07\%) & (5.48\%) & (\underline{11.09\%}) & (11.57\%) & (5.37\%) & (\underline{8.64\%}) & (7.75\%) & (6.82\%) & (\underline{8.66\%}) & (8.00\%) \\
\midrule
\multirow{2}*{8} & 49.83 & 52.66 & 51.97 & 118.82 & 122.16 & 121.79 & 80.02 & 83.63 & 83.15 & 36.52 & 43.88 & 43.01 & 70.31 & 72.72 & 72.75 & 42.09 & 50.25 & 49.47 \\
 & (11.41\%) & (12.72\%) & (9.66\%) & (4.25\%) & (4.41\%) & (3.76\%) & (4.27\%) & (4.94\%) & (3.56\%) & (10.25\%) & (10.17\%) & (6.88\%) & (5.25\%) & (8.81\%) & (6.44\%) & (6.27\%) & (8.71\%) & (7.51\%) \\
\bottomrule
\end{tabular}}
\caption{Results of \textit{SabER} under different $N$-$n$ settings across six datasets, where $N$ is the training sequence shot and $n$ is the generation sequence shot. VizWiz and MSCOCO are selected as representative datasets for the VQA and Captioning tasks. The data in the upper part of each cell shows \textit{SabER}'s performance, while the numbers in parentheses below indicate the improvement from task-aware attention. The data \underline{underlined} correspond to the setting in main experiments, i.e., 4-4.\label{shottable}}
\end{table*}
\label{shot}
Table \ref{shottable} shows that in all $N$-$n$ settings, including interpolation and extrapolation, task-aware attention in \textit{SabER} has a positive effect. \textit{SabER} achieves notably strong performance in the 4-8 setting, indicating its potential in both low-data scenarios and in ICL with more shots, even in many-shot ICL, as the context size of LVLMs increases. Overall, when training and generation shots are consistent, performance is maximized, as the task semantics learned by the model can be applied equally and evenly to guide sequence generation.

However, in the 8-8 setting, some performance metrics are unexpectedly lower than those in the 8-4 or even 4-4 settings. Given that LVLMs can perform better TL with more ICDs, this suggests that TR driven by task semantics plays a more significant role. We deduce that the task semantics in ICL exhibits marginal effects related to the number of ICD shots. This marginal effect accumulates through the task representation learned by \textit{SabER} via task-aware attention, and the task patterns recognized by the LVLM during TR from ICDs. 

Therefore, for tasks like VQA and \textbf{CLEVR}, balancing the varying TR dependence in well-trained LVLMs with the impact of task semantics during the training of configuration models is demanding. This highlights the importance of task-aware attention in flexible ICD sequence configuration. \textit{SabER} enables high-precision multimodal ICL tailored to specific needs.
\subsection{Input Embedding}
\label{embapp}
For the CLIP encoders, we explore three alternative methods: one involves freezing its parameters and adding an MLP adapter to its output, which is then trained; another involves fully training the entire encoder; and the third involves training only the last two layers. For constructing the embeddings multimodal ICD tokens, we first experimented with direct concatenation without gating modules:
\begin{equation*}
e_{i}=E_{I}(I_{i})+E_{T}(Q)_{i}+E_{T}(R_{i})+r_{i},
\end{equation*}
where $r_{i}$ is a randomly initialized learnable component introduced into the embedding. Besides binary gating, we examine a finer-grained ternary gating module that assigns separate weights to control the contributions of all three components $I$, $Q$ and $R$:
\begin{equation*}
   e_{i}=g_{I}\cdot E_{I}(I_{i}) +g_{Q}\cdot E_{T}(Q_{i})+g_{R}\cdot E_{T}(R_{i}),
\end{equation*}
where $g_{I}, g_{Q}$ and $g_{R}$ denote the weights computed using a softmax function applied the linear transformations, ensuring their sum equals 1. Additionally, we apply regularization to the weights: $g_{I}^2+g_{Q}^2+g_{R}^2\le \theta$ to prevent excessive reliance on specific components. 
\subsection{Generalization Test}
\begin{table}
\centering
\begin{tabular}{c c c c c c}
\toprule
\textbf{Datasets} & \textbf{Training} & \textbf{Validation} & \textbf{Test} & \textbf{$\hat{D}$ Size} & metrics\\
\midrule
Rule Learning & 1600 & - & 150& & exact match scores\\
Fast Counting & 800 & - & 40& & Accuracy\\
\bottomrule
\end{tabular}
\caption{\label{size1} Overview of Rule Learning and Fast Counting tasks.}
\end{table}

For NLP evaluation, we utilize the Rule Learning part of the latest benchmark, ICLEval. ICLEval is designed to assess the ICL abilities of LLMs, focusing on two main sub-abilities: exact copying and rule learning. The Rule Learning part evaluates how well LLMs can derive and apply rules from examples in the context. This includes tasks such as format learning, where models must replicate and adapt formats from given examples, and order and statistics-based rule learning, where the model must discern and implement patterns such as item sequencing or handling duplications. These tasks challenge LLMs to go beyond language fluency, testing their ability to generalize from context in diverse scenarios. Examples of $(Q,R)$ pairs can be found in Table \ref{icleval}. For all tasks, we use exact match scores to evaluate the predictions with the labels.

For text-to-image evaluation, we utilize the Fast Counting task in the VL-ICL bench. In this task, artificial names are associated with the counts of objects in the image. The task is to generate an image that shows a given object in quantity associated with the keyword (e.g. perpo dogs where perpo means two). Thus, each $Q$ is a two-word phrase such as 'perpo dogs', and its corresponding $R$ is an image of two dogs.

\begin{table*}

\centering
\begin{tabular}{m{2cm}<{\centering} | m{6cm}| m{4cm}}
\toprule
\textbf{Task} & \textbf{$Q$} &\textbf{$R$}\\
\midrule
Format rules & |Index|name|age|city|\newline
|---|---|---|---|\newline
|1|Elijah Morgan|36|Pittsburgh| & $<$person>\newline
$<$name>Elijah Morgan$<$/name>\newline $<$age>36$<$/age> \newline$<$city>Pittsburgh$<$/city>\newline
$<$/person>\\

\midrule
Statistics rules & 588 and 823 are friends.\newline
885 and 823 are friends. \newline
795 and 588 are friends.\newline
890 and 823 are friends.\newline
885 and 588 are friends.\newline
890 and 588 are friends.\newline
795 and 823 are friends.\newline
Query: Who are the friends of 885? & 823, 588 \\

\midrule
Order rules& Input: activity, brief, wonder, anger\newline Output: anger, wonder, activity, brief \newline Input: market, forever, will, curve \newline Output: curve, will, market, forever \newline Input: pain, leading, drag, shoot \newline Output: shoot, drag, pain, leading \newline Input: shopping, drama, care, start \newline  Output: &start, care, shopping, drama\\

\midrule
List Mapping & Input: [1, 3, 6, 1, 83]\newline
Output: [3]\newline
Input: [5, 6, 35, 3, 67, 41, 27, 82]\newline Output: [6, 35, 3, 67, 41]\newline
Input: [8, 45, 6, 18, 94, 0, 1, 2, 7, 34]\newline Output: [45, 6, 18, 94, 0, 1, 2, 7]\newline Input: [2, 7, 66, 6, 93, 4, 47]\newline
Output: & [7, 66] \\

\bottomrule

\end{tabular}
\caption{\label{icleval} The examples of four Rule Learning tasks in ICLEval. }

\end{table*}

\subsection{ICD Sequence Evaluation}
\begin{table}
    \centering
    \begin{tabular}{lcccccccc}
        \toprule
        \multirow{2}{*}{\textbf{Method}} & \multicolumn{2}{c}{VQAv2} & \multicolumn{2}{c}{VizWiz} & \multicolumn{2}{c}{OK-VQA} & \multicolumn{2}{c}{\textbf{Hybrid}} \\
        \cmidrule(lr){2-3} \cmidrule(lr){4-5} \cmidrule(lr){6-7} \cmidrule(lr){8-9}
        & Gap ↑& Variance↓ & Gap↑ & Variance↓ & Gap↑ & Variance↓ & Gap↑ & Variance↓ \\
        \midrule
        \textbf{I2I} & 2.86 & 22.61 & 1.83 & 25.34 & 3.07 & 21.94 & 1.54 & 26.79 \\
        \textbf{IQ2IQ} & 3.27 & 21.96 & 2.79 & 26.57 & 3.43 & 19.51 & 2.31 & 25.34 \\
        Lever-LM & 3.42 & 16.21 & 3.64& 18.57 & 3.08  & 18.18 & 2.76& 20.85 \\
        Ours & \textbf{3.85} & \textbf{14.82} & \textbf{3.85} & \textbf{16.34} & \textbf{3.37} & \textbf{13.77} & \textbf{3.39} &  \textbf{17.98}\\
        \bottomrule
    \end{tabular}
    \caption{Results of ICD sequence evaluation of four configuration methods. The best scores are highlighted in \textbf{bold}.}
    \label{seq}
\end{table}

Based on our understanding of the ICD sequences in Section \ref{rethink}, we conduct experiments on four datasets where the short-cut effect is the most prevalent. To evaluate the average quality of ICD sequences, we use two metrics: Gap, which measures the average performance difference after randomly replacing one ICD in a sequence with another ICD from the same sequence (resulting in one ICD being duplicated), and Variance, which quantifies the variance in sequence performance for a given configuration method on a specific task. The results are presented in Table \ref{seq}. \textit{SabER} achieves the highest Gap across all four datasets, indicating that the ICD sequences it constructs exhibit a more comprehensive task mapping. Additionally, it consistently demonstrates the lowest Variance, suggesting that the task mappings within its sequences are the most accurate and stable, minimizing reliance on shortcut inference. 
\end{document}

%% file: math_commands.tex
\usepackage{amsmath,amsfonts,bm}

\def\eqref#1{equation~\ref{#1}}
\def\1{\bm{1}}

\DeclareMathAlphabet{\mathsfit}{\encodingdefault}{\sfdefault}{m}{sl}
\SetMathAlphabet{\mathsfit}{bold}{\encodingdefault}{\sfdefault}{bx}{n}